\documentclass[journal,twoside,web]{ieeecolor}
\usepackage{generic}
\usepackage{cite}
\usepackage{amsmath,amssymb,amsfonts}
\usepackage{algorithmic}
\usepackage{graphicx}
\usepackage{textcomp}
\usepackage{soul}
\usepackage{url}
\usepackage{multirow}
\usepackage{colortbl}
\def\BibTeX{{\rm B\kern-.05em{\sc i\kern-.025em b}\kern-.08em
    T\kern-.1667em\lower.7ex\hbox{E}\kern-.125emX}}
\begin{document}
\title{Machine Learning Regression based Single Event Transient Modeling Method for Circuit-Level Simulation} 
\author{ChangQing Xu, \IEEEmembership{Member, IEEE}, Yi Liu, XinFang Liao, JiaLiang Cheng and YinTang Yang, \IEEEmembership{Member, IEEE}
\thanks{ ``This work was supported in part by the Fundamental Research Funds for the Central Universities, by the Industry-University-Academy Cooperation Program of Xidian University-Chongqing IC Innovation Research Institute under Grant CQIRI-2021CXY-Z01, by Equipment Pre-research Project of China under Grant 41424050607.'' }
\thanks{Changqing Xu,Yi Liu, XinFang Liao, JiaLiang Cheng, YinTang Yangon, are with the School of Microelectronics, Xidian University, Xi’an 710071, China (e-mail: cqxu@xidian.edu.cn).}
}
\maketitle 

\begin{abstract}
In this paper, a novel machine learning regression based single event transient (SET) modeling method is proposed. The proposed method can obtain a reasonable and accurate model without considering the complex physical mechanism.  We got plenty of SET current data of SMIC 130nm bulk CMOS by TCAD simulation under different conditions (e.g. different LET and different drain bias voltage). A multilayer feedfordward neural network is used to build the SET pulse current model by learning the data from TCAD simulation. The proposed model is validated with the simulation results from TCAD simulation. The trained SET pulse current model is implemented as a Verilog-A current source in the Cadence Spectre circuit simulator and an inverter with five fan-outs is used to show the practicability and reasonableness of the proposed SET pulse current model for circuit-level single-event effect (SEE) simulation.
\end{abstract}


\section{Introduction}
\label{sec:introduction}
\IEEEPARstart{W}{ith} continuous downscaling of transistor sizes and supply voltages, the sensitivity to single event transient (SET) has become one of the most important reliability issues for aerospace integrated circuit (IC) which is under radiation exposure \cite{a1,a2,a3}. Generally, the evaluation and analysis of the circuit's sensitivity to SETs can be performed either by the computer-aided modeling and simulations or through the experimental approach. As the irradiation experiments are very expensive, circuit-level single event transient (SET) simulation is suitable for understanding how the circuit responds to the ion strike and how the circuit is hardened to avoid SET at the early stages of the design \cite{a6}. SET model has been widely applied in analogy and digital circuit SET simulation\cite{a4,a5,a6}, which can guide radiation-hardened IC design.
\begin{figure}[t]
\centerline{\includegraphics[width=\columnwidth]{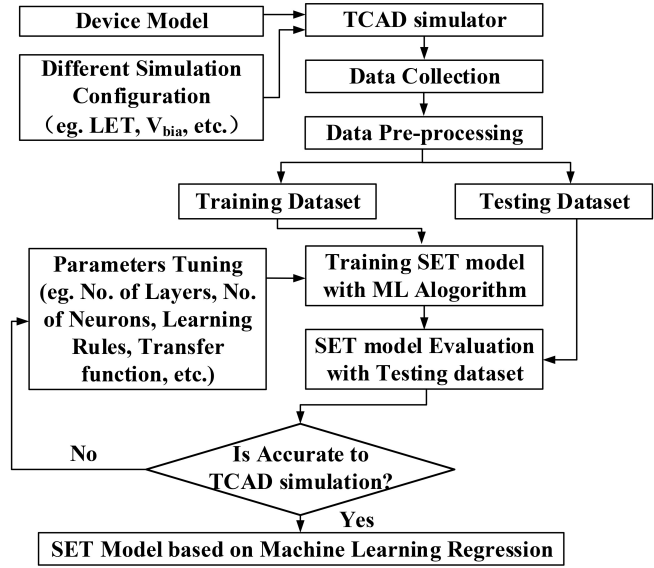}}
\caption{Flow of the SET modeling method based on Machine Learning Regression.}
\label{fig1}
\end{figure}

However, the accuracy of the SET model have a direct impact on circuit-level SET simulation.To obtain an accurate SET model, lots of researchers proposed various approaches to build SET model \cite{a7,a8,a9,a10,a11}. A double-exponential model \cite{a2,a7} is widely used to model the SET current from the drain to source of NMOS. Another exponential model is proposed by Freeman \cite{a8}, in which a single technology-related timing parameter is defined to make the mode suitable for different technologies.
Hu's current model considering the angel of particle incidence \cite{a9}.
Considering the influence of strike location, a bias-dependent single event model is proposed in which  a drift factor and a diffusion factor is introduced to produce the accurate single-event current pulse \cite{a11}. 
There are also some physics-based SET models \cite{a12,a13}, which can transient potential and drain current on linear energy transfers (LETs), strike positions, drain and gate biases, etc. Physics-based SET can capture the essential physics behind these current transients, but these models are too complex to use in circuit-level simulation. Those SET models, which are based on certain functions, are suitable for circuit-level simulation. However, due to the complex physical mechanism of devices and single event effect, it is hard to build an accurate SET current model based on a certain function, such as double-exponential function, when various variables, such as LET, strike location, bias voltage, etc, are considered. 

In this paper, we proposed a novel SET modeling method based on machine learning regression technology. A feedfordward neural network is used to train the SET pulse current model. The trained model is validated and implemented as a Verilog-A current source in the Cadence Spectre circuit simulator.

\section{Machine Learning Regression based SET Modeling Method}
Fig. \ref{fig1} shows the flow of SET modeling method based on machine learning regression where all steps have been partition into phases. The Cogenda VisualTCAD\cite{a14}, which consists of Gds2mesh, VisualParticle, and genius, is used to generate a dataset for training and testing the SET pulse current model. Simulation results are collected over various LET, drain bias voltage, and simulation time. These data will be pre-processed and partitioned as training and testing datasets. A multi-layer feedforward neural network is applied as SET model and trained based on the dataset obtained from TCAD simulation. To make a trade-off between the accuracy and the complexity of the SET model, the parameters, such as neural network layers, number of neurons, transfer functions, etc, are tuned based on the accuracy requirement.

\subsection{SET Simulation in TCAD and Data Collection}
\begin{figure}[t]
\centerline{\includegraphics[width=\columnwidth]{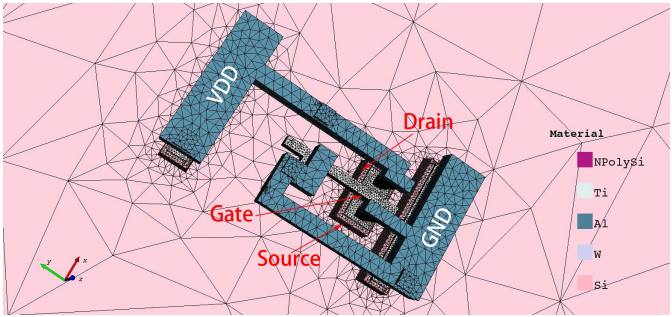}}
\caption{3D  structure  of  the  NMOS.}
\label{fig2}
\end{figure}
In this paper, all TCAD simulation results are obtained by Cogenda VisualTCAD\cite{a14}. 
Fig. \ref{fig2} shows the 3D structure of the NMOS in a Semiconductor Manufacturing International Corporation (SMIC) 130nm CMOS technology.
The device dimension of NMOS for the model and  TCAD simualtions in the following sections are channel length $L$ = 130 nm, channel width ${W}$ = 420 nm, gate oxide thinkness ${t_{ox}}$ = 2.58 nm, source/darin doping is $10^{20} cm^{-3}$, and channel doping $N = 1.57\times10^{17}cm^{-3}$. 
The gate voltage is fixed at 0 V, source and substrate are grounded and the drain voltage is set to $0 \sim 1.8 V$.

The heavy ion trike is simualted using heavy ion model in Cogenda VisualTCAD \cite{a14}. The models used in the simulation are Shockley-Read-Hall, Auger, and direct (or radiative) recombination. SRH carrier lifetime has both doping dependence and temperature dependence. Other models applied are effective density of states model, Fermi–Dirac statistics, impact ionization, and Lucent mobility model. Heavy ion samples with LET of $4 \sim 100 MeV \cdot cm^2/mg$, The incident positions of these ions were the NMOS drain center, the track length of the heavy ions was set to be 10 $\mu m$, and the incident direction is the direction vertical to the surface of the device.
TCAD simulation was conducted for SMIC 130nm NMOS and the drain current data at various drain bias voltages($0 \sim 1.8 V$) and various LETs($4 \sim 100 MeV \cdot cm^2/mg$) is collected. LET, drain bias voltage and time have been considered as input features and drain current has been divided as output features. Because the data obtained in TCAD simulation is too dense at the corner of the curve which may cause neural network over fitting, cubic spline data interpolation is used to get the sampling point based on accuracy requirements. 
418000 data is obtained based on TCAD simulation results.
Among these data, 70\% of data will be selected randomly and considered as training data, 15\% data will divided into validation and remaining 15\% data will be set as test data.  

\subsection{SET pulse current model based on feedfordword neural network}
In this paper, feedforward neural networks, which consist of an input layer, several hidden layers, and an output layer, are used to build SET pulse current model. Levenberg–Marquardt algorithm is used as learning rule. Because the number of neurons, number of layers and transfer function applied will directly affect the fitting ability of the neural network. Several neural networks are trained 1000 epoch to build SET pulse current model and the mean square errors(MSE) are shown in Table \ref{table1}.  Considering the complexity and accuracy of the SET model, the neural network(8$\times$8$\times$1) is selected to build SET pulse current model.

\begin{table}[t]
\caption{MSE of different neural networks after training}
\label{table1}
\setlength{\tabcolsep}{3pt}
\centering
\begin{tabular}{llll} 
\hline
\multirow{2}{*}{Network}                & \multicolumn{2}{c}{Tranfer Function} & \multirow{2}{*}{Error
(MSE)}  \\ 
\cline{2-3}
                                        & Hidden
Layer & Output
Layer          &                               \\ 
\hline
16$\times$1                                    & tansig       & purelin               & 9.54e-3                             \\
16$\times$1                                    & logsig       & purelin               & 2.48e-2                             \\
16$\times$1                                    & elliotsig       & purelin               & 1.83e-2                            \\
\rowcolor[rgb]{0.737,0.722,0.722} 8$\times$8$\times$1 & tansig       & purelin               & 1.37e-3    \\
8$\times$8$\times$1 & logsig       & purelin               & 2.14e-3   \\
8$\times$8$\times$1 & elliotsig       & purelin               & 2.18e-2  \\
8$\times$16$\times$8$\times$1                                 & tansig       & purelin               & 2.76e-4                             \\
8$\times$16$\times$8$\times$1                                 & logsig       & purelin               & 4.99e-4                             \\
8$\times$16$\times$8$\times$1                                & elliotsig       & purelin               & 2.10e-2                             \\
\hline
\end{tabular}
\end{table}

\begin{figure}[t]
\centerline{\includegraphics[width=\columnwidth]{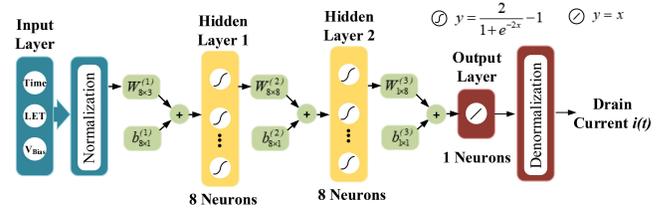}}
\caption{SET model based on feedfordward neural network}
\label{fig4}
\end{figure}
Fig. \ref{fig4} shows SET pulse current model based on 8$\times$8$\times$1 feedfordward neural network. In the input layer, time, LET and bias voltage are the input variables. In the hidden layers, tansig function is used as transfer function. In the output layer, a simple line function is selected as transfer function. $W^{(1)}_{8\times3}$, $W^{(2)}_{8\times8}$, $W^{(3)}_{1\times8}$ are trained weight matrices for corresponding layers. $b^{(1)}_{8\times1}$, $b^{(2)}_{8\times1}$, $b^{(3)}_{1\times1}$ are trained bias matrices for corresponding layers.
Due to different input data units, all input data need to be normalized first and the neural network output will be denormalized to generate the accurate SET pulse current. 

\section{Results and Discussion}
\begin{figure}[t]
\centerline{\includegraphics[width=\columnwidth]{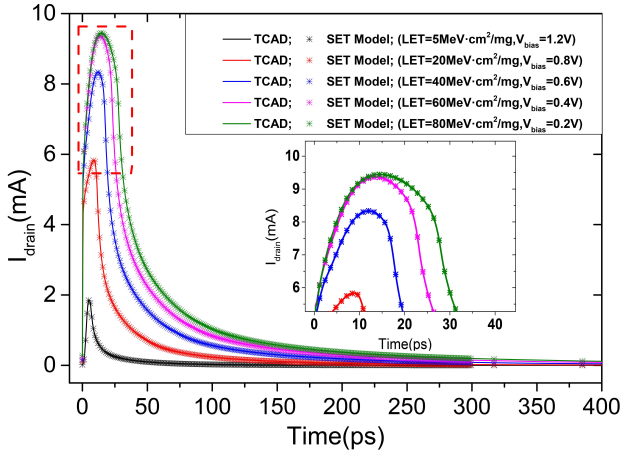}}
\caption{SET current versus time for different drain voltage and LET}
\label{fig5}
\end{figure}
\begin{figure}[t]
\centerline{\includegraphics[width=\columnwidth]{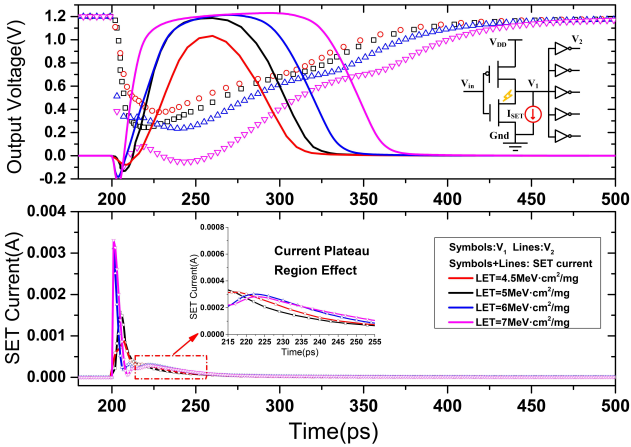}}
\caption{SET current and output voltage of the two-stage inverter versus time for different LET }
\label{fig6}
\end{figure}
To show the accuracy of the trained SET pulse current model, we compare trained model and TCAD simulation results. Fig. \ref{fig5} shows the SET pulse current model under various drain bias voltage (0.2, 0.4, 0.6, 0.8, 1.2V) and various LET (5, 20, 40, 60, 80 $MeV \cdot cm^{2}/mg$) with respect to time and compared with TCAD simulation. As Fig. \ref{fig5} is shown, the trained model can fit the SET pulse current very well, especially for the peak value and width of current pulse which is of great importance to circuit-level simulation.

The developed SET pulse current model is implemented as a Verilog-A current source $I_{SET}$ in the Cadence Spectre circuit simulator. The SET simulation is carried out in the two-stage CMOS inverter by connecting the SET current source in parallel to the OFF-state NMOS in the first-stage of inverter as shows in Fig. \ref{fig6}. The first-stage inverter, which is five fin-outs, is connected to five inverters. The ion strike at the OFF-state NMOS in the first-stage inverter and ion strike time is 200ps. As shown in Fig. \ref{fig6}, the ion strike at the OFF-state NMOS in the first-stage inverter at 200 ps, which causes the output voltage is pulled down from the logic high state. A plateau, which is caused by the dynamic interaction of the depressed drain voltage due to the strike and the compensating PMOS transistor drive current \cite{a15,a16}, can be observed in the SET current of nMOS in the inverter. For higher LETs, the voltage perturbation is more obvious due to higher amount of charge deposition in the silicon body.

\section{Conclusion}
A novel machine learning regression based SET modeling method is proposed and a SET pulse current model based on feedforward neural network is developed and validated.
The developed model is easy to integrate into the circuit simulator and takes lesser computation time compared to the TCAD simulations. The analysis can be extended to other digital or analog circuits using the proposed SET pulse current model, which helps the circuit designers to predict and mitigate the influence of SET on circuit design.




\end{document}